\documentclass[letterpaper, 10pt, twocolumn]{article}
\usepackage{clean}

\usepackage{cite}
\usepackage{amsmath,amssymb,amsfonts}
\usepackage{algorithmic}
\usepackage{graphicx}
\usepackage{textcomp}
\usepackage[colorlinks=true, linkcolor=black, urlcolor=cyan, filecolor=black, citecolor=black]{hyperref}
\usepackage{xcolor}
\def\BibTeX{{\rm B\kern-.05em{\sc i\kern-.025em b}\kern-.08em
    T\kern-.1667em\lower.7ex\hbox{E}\kern-.125emX}}

\begin{document}

\title{Industrial Cabling in Constrained Environments: a Practical Approach and Current Challenges
\thanks{This project has received funding from the European Union's Horizon 2020 research and innovation programme under grant agreement No 101058521 — CONVERGING.}
}
\author{Tanureza Jaya,~Benjamin Michalak,~Marcel Radke,~Kevin Haninger\thanks{All authors with Fraunhofer IPK, Berlin, Germany. Email: \texttt{fistname.lastname@ipk.fraunhofer.de}}}

\maketitle

\begin{abstract}
Cabling tasks (pulling, clipping, and plug insertion) are today mostly manual work, limiting the cost-effectiveness of electrification. Feasibility for the robotic grasping and insertion of plugs, as well as the manipulation of cables, have been shown in research settings. However, in many industrial tasks the complete process from picking, insertion, routing, and validation must be solved with one system. This often means the cable must be directly manipulated for routing, and the plug must be manipulated for insertion, often in cluttered environments with tight space constraints.

Here we introduce an analysis of the complete industrial cabling tasks and demonstrate a solution from grasp, plug insertion, clipping, and final plug insertion. Industrial requirements are summarized, considering the space limitations, tolerances, and possible ways that the cabling process can be integrated into the production process.

This paper proposes gripper designs and general robotic assembly methods for the widely used FASTON and a cubical industrial connector. The proposed methods cover the cable gripping, handling, routing, and inserting processes of the connector. Customized grippers are designed to ensure the reliable gripping of the plugs and the pulling and manipulation of the cable segments. A passive component to correct the cable orientation is proposed, allowing the robot to re-grip the plug before insertion. In general, the proposed method can perform cable assembly with mere position control, foregoing complex control approaches. 

This solution is demonstrated with an industrial product with realistic space requirements and tolerances, identifying difficult aspects of current cabling scenarios and potential to improve the automation-friendliness in the product design.
\end{abstract}

\section{Introduction}
Due to the prevalent use of electronic components in various industrial branches, cable assembly has become a common task in the production process. However, cable assembly still proves to be a challenging process to automate due to numerous reasons \cite{trommnau}. The high variance of connector designs indicates that certain plugs require specialized gripper fingers. Furthermore, certain cable plugs have tighter tolerances (0.3-0.6 mm \cite{yumbla}) that requires accurate positioning for a reliable insertion \cite{yumbla2} or misalignment compensation methods \cite{hartisch}. 

The flexibility of the cable results in inconsistent position and orientation of the plug, which complicates the gripping as well as insertion processes \cite{navas}. Cables may require complex routing in a tight workspace, limiting the freedom of movement for a possible robotic system as well as restricting the finger design of the gripper. 

This paper proposes gripper designs and general robotic assembly methods for two types of connectors in a realistic industrial environment. The proposed methods cover the cable gripping, handling, routing, and insertion processes of the connectors, and the results can be seen at \url{https://youtu.be/cv5sMI0DYxM}.

\section{State of Technology}
In general, cable assembly task consists of three steps: cable gripping, routing, and plug insertion. This chapter discusses established methods and relevant aspects for each steps. While there are many innovative methods regarding cabling assembly, many of them are introduced in simplified applications, which have significant gaps in tolerances and space requirements compared to realistic industrial applications. 

\subsection{Cable Gripping}
Robotic cable assembly starts with the robot picking the cable, which can be stored in multiple ways. The cable plug can be fixed to a magazine to ensure a consistent position and orientation, allowing the plug to be grasped with just position control. The cable can also be loosely stored with indefinite position and orientation. In such cases, the robot can implement a camera system to identify the cable location and orientation before picking it \cite{haraguchi}. When the plug is mounted on the cable at the robot station (in situ assembly), the cable preparation machine can also act as a feeder to supply the cable \cite{trommnau}. 

Alternatively, the robot gripper can be equipped with an actuator system that can manipulate the cable after it has been picked. For example, a gripper can be equipped with a vibration plate \cite{yumbla} or a roller system \cite{yumbla3} to manipulate the cable alignment within the robot's grip. In \cite{zhou}, numerous additional actuators and sensors are integrated to the gripper to allow twisting and sliding of the cable, as well as identifying the cable direction. However, additional actuators increase the size of the gripper quite considerably, which puts its viability for tight workspaces in question.

\subsection{Cable Routing}
While the plugs on both cable ends are inserted to the sockets, the cable segment is secured through routing points connected to the workpiece. While cable routing can be performed by a robot with a gripper, there are cable routing methods that involve multiple cable manipulators. In \cite{heisler}, the robot flange is equipped with multiple grippers, whereby one gripper holds the cable plug while another gripper pushes the cable into the routing fork. And in \cite{zhu}, a second robot arm is implemented to offer even more freedom in handling the cable. 

Beside securing the cable segment to the routing points, the cable shape is also an important aspect of the routing process. The cable must be manipulated in such a way, so that the cable can fit to the available space among other components and avoid getting stuck with the environment \cite{jiao}. In this regard, several frameworks have been introduced to manipulate the cable shape through the robot movement \cite{waltersson, zhang}.

\subsection{Plug Insertion}
A prerequisite for a successful plug insertion is that the gripped plug must align with the socket. Assuming the robot accuracy is high enough and the workpiece position is fixed, e.g. through clamping, the robot can find the socket simply by using position control. Otherwise, the alignment of the plug and the socket must be ensured with another method. 

Camera systems can be implemented to aid the insertion process, whereby two cameras that are positioned perpendicular to each other to observe position deviation of the plug from two different angles and correct the robot pose accordingly \cite{sun}. Nonetheless, this method requires a clear view of the plug and socket and will be rendered ineffective in a realistic environment, where the view of the socket is obstructed by other components. 

Furthermore, search strategies, such as spiral and probing search are proposed in \cite{haraguchi} to find the socket position. However, these search strategies require the robot to move freely around the socket, which is unattainable in tight workspaces.

\section{Setup of Cable Assembly}
This chapter discusses the cable assembly performed in this paper. The robot Franka Research 3 from the manufacturer Franka Emika, which has seven joints with integrated torque sensors and a parallel gripper, is utilized for the cable assembly.

\begin{figure}
    \centering
    \includegraphics[width=1\linewidth]{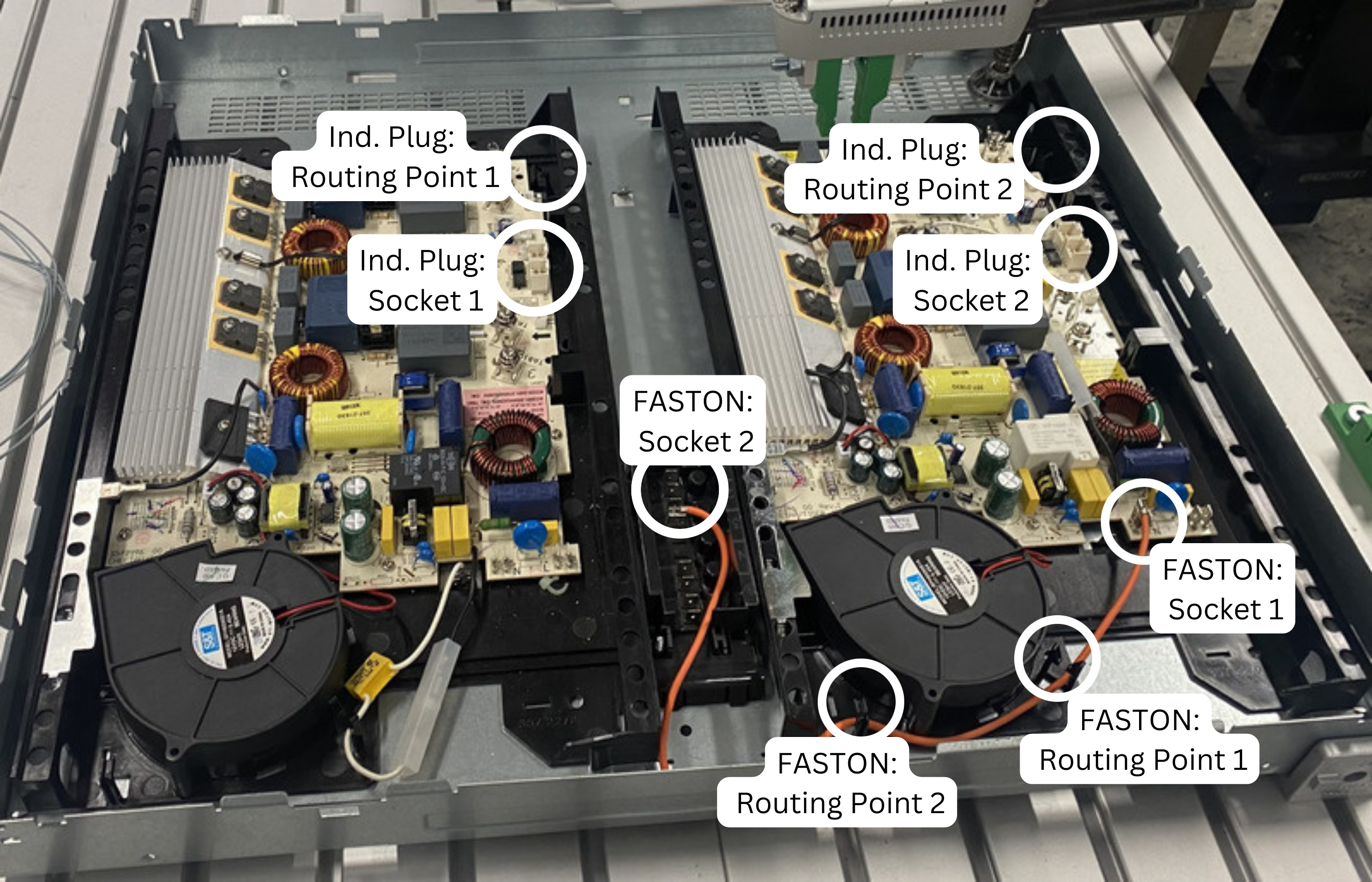}
    \caption{Workspace of Cable Assembly}
    \label{fig:enter-label}
\end{figure}

The cable assembly is performed on a metal board with electronics using two types of connectors; FASTON connectors with single thread cable and another industrial connector with multiple cables. To ensure the position of the sockets on the metal board is consistent, the metal board is clamped. Therefore, all inaccuracies during the assembly process originate from the robot end effector, which can amount to 1-2mm.

The workspace of this assembly corresponds to a realistic industrial work environment. Many other components are compactly installed on the metal board. Therefore, there is limited space for the robot movement on the area around the socket and the routing points. 

Typically, the insertion direction of a plug is parallel to the connected cable \cite{yumbla2, hartisch}. In this paper, the insertion direction of both connector types are perpendicular to the cable. As a result, the plugs are inserted to the socket vertically, needing grasping solutions which allow sufficient force in this direction.

Two types of magazine are designed for the FASTON and the industrial connector, which holds the plugs while leaving the contact area of the plug free for the gripper. This allows the robot to pick the connectors from above the magazine. Due to the varying constraints, two different cabling methods are proposed for the two connector types. 

\section{Cable Assembly: FASTON}
The proposed cabling method for the FASTON connector is depicted in the following table. All of the assembly steps are performed using mostly position control. The magazine for the FASTON connector only holds the plug on one end of the cable, while the second plug is left free.

\begin{table}[h]
\caption{Assembly Method for FASTON connectors}
\begin{tabular*}{\hsize}{@{\extracolsep{\fill}}lp{8cm}@{}}
\hline
Step & Description \\
\hline
1 & The robot moves above the magazine with an open gripper. Afterwards, it moves downwards and grips the first end of the plug.\\
2 & The robot moves above the first socket and moves downwards to insert the plug.\\
3 & The robot grips and routes the cable.\\
4 & The robot grips the second plug with the help of a orientation correction device.\\
5 & The robot moves above the second socket and moves downwards to insert the plug.\\
\hline
\end{tabular*}
\end{table}

\subsection{FASTON Gripper}
In the proposed cabling method for the FASTON plug, the robot grips the plug in the insertion process, and the cable in cabling process. Therefore, a finger design that is suitable for the gripping of both plug and and cable is required. 

Due to the round and asymmetrical form of the FASTON plug, a standard finger with a flat surface can not grip it securely. Instead, a form-fit finger design is implemented with a finger tip cutout that fits the geometry of the FASTON plug. 

In order to create such cutout, the finger is fabricated using a 3D printer. The cutout is designed to also fit the cable diameter, therefore the finger can grip the rubber cable segment reliably. On the other hand, it can not grip the plug, which is made out of metal, as tightly due to the lower friction coefficient.

\begin{figure}
    \centering
    \includegraphics[width=0.5\linewidth]{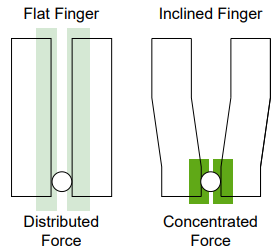}
    \includegraphics[width=0.5\linewidth]{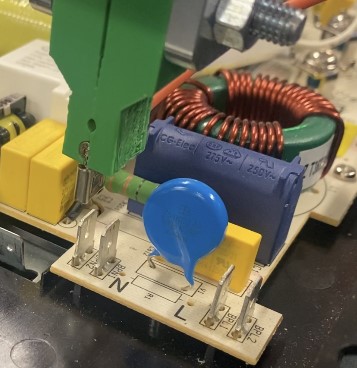}
    \caption{Finger Design for FASTON Connectors}
    \label{fig:enter-label}
\end{figure}

Usually the finger surface is layered with rubber to achieve higher friction coefficient. However, applying additional rubber layer to such a small cutout will negate the form-fit design of the fingertip. Instead, the tip of the finger is designed with an inward incline. With this incline, the force generated by the gripper is concentrated on the fingertip instead of distributed along the whole finger. With this design, the printed finger is able to firmly grip the metal plug without the help of a rubber layer. Furthermore, the fingertip is designed with a narrow geometry to comply with the limited space around the sockets and routing points.

\subsection{Cable Routing}
After the first plug is inserted, the robot releases the plug and loosely grips the the part of the cable that is located right beside the first plug to start the routing process. The robot can reliably find this part of the cable, since its position can be derived from the position of the attached plug.

While gripping the cable, the robot can open the gripper slightly to loosen the grip without dropping the cable. Since one end of the cable is fixed to the first socket, the robot can slide the gripper finger along the cable in the direction of the loose second plug.

With the slightly opened gripper the robot moves toward the first routing point, while the gripper slides along the cable. After reaching the the routing point, the robot grips the cable firmly. In this state, the robot can insert this particular cable segment to the routing point.
Afterwards, the robot again loosens the grip on the cable and moves to the following routing point. The same process is then repeated to route the next cable segment.

This method allows the robot to reach different cable segments by moving the gripper along the cable using a loose grip. It also allows the robot to manipulate and insert a particular cable segment to the corresponding routing point usig a tight grip.

The routing process of the FASTON connector is carried out in a constrained area. The first routing point is located right beside another component, while the second routing point is located between a component and the edge of the metal board. As a result, the robot can not reach the routing points directly, and the cable must be inserted to the routing points from a specific direction. While the maneuverability of the robot around the routing points is improved by the narrow fingertip design, it still requires numerous auxiliary positions to move around the routing points without colliding with the immediate environment.

\subsection{Cable Orientation Correction}
After the routing process, the robot moves the gripper along the cable until it reaches the second end of the cable, right before the second plug. At this point, the robot aims to grip and insert the second plug. However, the orientation of the second plug is still undetermined. 

\begin{figure}
    \centering
    \includegraphics[width=1\linewidth]{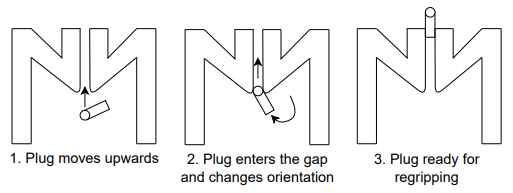}
    \caption{Alignment Method for FASTON Plugs}
    \label{fig:orientation}
\end{figure}

This paper introduces an orientation correction component that aims to bring the second plug to an upright orientation to enable a reliable grip from the robot. This component, which is installed beside the workpiece, consists of two pillars with a gap between them (figure \ref{fig:orientation}). 

When the robot tightly grips the cable, it can not move in the axial direction, but can still rotate, hence the plug can also revolve. The introduced component utilizes this property to correct the orientation of the plug.

Afterwards, the robot moves to a position, whereby the plug is located underneath the gap. The robot will then move upwards with the plug. As the plug enters the gap, the curved surface around the gap will force the plug to an upright orientation. The robot will stop moving upwards and release the cable once the upper half of the plug is above the gap. The gap is designed to be narrower than the width of the plug, so that the tension from the pillars will keep the plug in the upright orientation even when the cable is released. Subsequently, the robot can reliably re-grip the plug for the second insertion.  

While the proposed component can reliably define the plug orientation in this use case, this correction method has its limitations. Firstly, the principle of the component only works with connectors with single cable thread, since this method relies on the cable to rotate while being gripped. Secondly, the method does not work in the exceptional case, whereby the plug faces exactly upwards as it enters the gap, 

\subsection{Plug Insertion}

\begin{figure}
    \centering
    \includegraphics[width=1\linewidth]{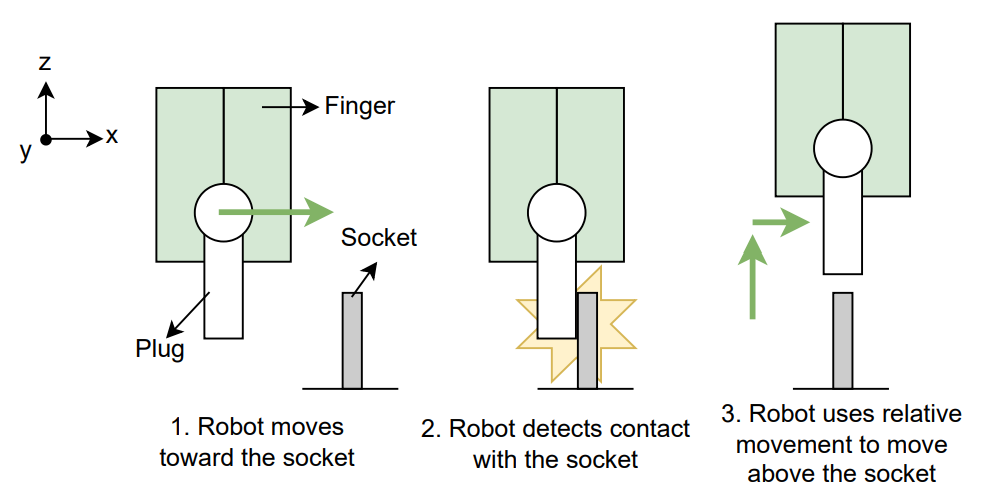}
    \caption{Contact-Based Insertion Method for FASTON Plugs}
    \label{fig:faston-insert}
\end{figure}

In the insertion process, each end of the plug is inserted into the socket vertically from above. The FASTON plug is relatively thin and therefore has a tight insertion tolerance (less than 1mm) on the x-axis in figure \ref{fig:faston-insert}. The sockets are also located in constrained areas. The first socket is located in front of other components, while the second socket is assembled in a small box (Figure 1). Furthermore, the sockets are assembled closely next to one another. 

To compensate for the tight insertion tolerance, while also complying to the space restrictions, an insertion method based on contact recognition is proposed. Instead of directly inserting the plug from above the socket, the robot starts the insertion process by moving to the side of the socket, with the broader side of the plug facing the socket. Afterwards, the robot moves horizontally toward the socket. The force sensor of the robot will detect when the plug comes into contact with the socket. Once a contact is detected, the robot stops. 

At this point, the plug is located right beside the socket. By moving the robot by a certain distance relative to where the contact is detected, the plug can be aligned to the socket on the x-axis. As depicted in step 3 of figure 4, the robot moves upward and then horizontally along the x axis. The distance of this horizontal movement shall be approximately the thickness of the plug.

Due to the chamfered edges of the socket, which is the male connector, and the innate compliance of the robot end-effector, the alignment tolerance on the y-axis is more relaxed and does not require said contact recognition method. Therefore, the plug can be inserted after the third step depicted in figure \ref{fig:faston-insert}. The robot inserts the plug by moving downward toward the socket. The proposed insertion method can still function even when the FASTON plug is placed with minimum distance (around 3mm) beside the socket. Therefore, it does not require a lot of space and thus is suitable in tight workspaces.

\section{Cable Assembly: Industrial Plug}

A different cabling method is proposed for the second connector type. Contrary to FASTON connectors, this type of industrial connector is connected to multiple cables and has a different geometry. Therefore, the proposed method to correct the orientation of a FASTON plug (chapter 4.3) can not be implemented on this industrial connector. 

Instead, the plugs on both cables ends are stored in the magazine. In this method, the robot to pick the second plug from the magazine after inserting the first plug. Afterward the robot can route the cable while gripping the second plug. Finally, the robot inserts the second plug after the routing process. 

In summary, the assembly method for the industrial plug is similar to the FASTON connector (Table 2), except the robot grips the second plug before the routing process.

\subsection{Industrial Plug Gripper}
A typical robot finger can be mounted on a parallel gripper to handle a single thread cable. In this case, the finger transfers the horizontal gripping force generated by the gripper, assuming the robot flange is upright. This gripping force is enough to grasp the single thread cable firmly. 

However, a wide range of industrial plugs are connected to multiple cables, which are aligned horizontally next to one another. Furthermore, a portion of such industrial plugs must be fully inserted into the socket. In this case, there is barely any accessible contact area for the gripper, hence it is not possible for the robot finger to grip the plug during the insertion process (figure 5). Instead, the robot must insert the plug while gripping the cable. 

\begin{figure}
    \centering
    \includegraphics[width=0.9\linewidth]{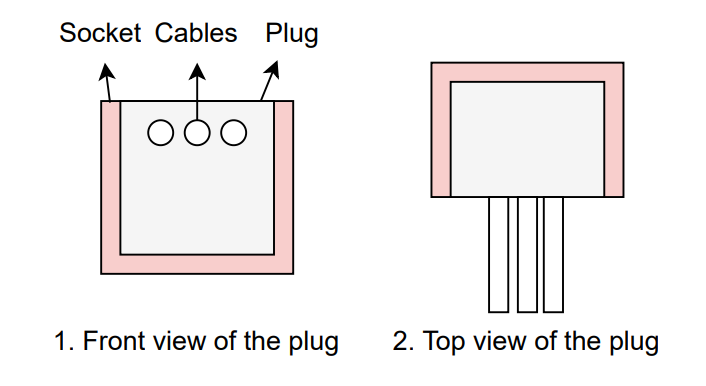}
    \caption{Unavailable Grip Surface on the Industrial Plug}
    \label{fig:enter-label}
\end{figure}

Nonetheless, a typical parallel gripper finger proves to be ineffective in handling such industrial plugs. While still assuming that the robot flange is upright, the horizontal force generated from a parallel gripper will work against the cables' alignment. If the robot flange is rotated to grasp the cables horizontally, collision with the workpiece will occur during insertion. 

\begin{figure}[htbp]
    \centering
    \includegraphics[width=0.6\linewidth]{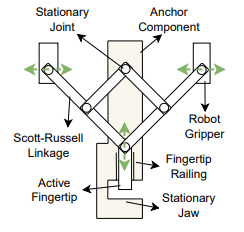}
    \includegraphics[width=0.5\linewidth]{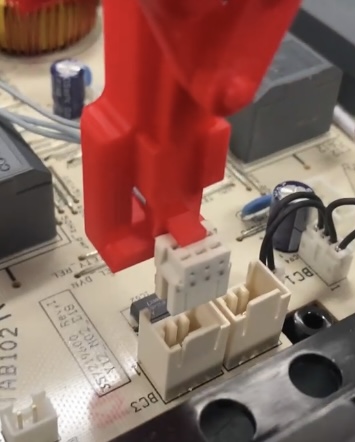}
    \caption{Finger Design for Industrial Plug}
    \label{fig:enter-label}
\end{figure}

The proposed solution for this problem is to grip the cables with a vertical gripping force to conform to the cable alignment, while maintaining an upright robot flange to avoid collision with the workpiece. This solution is achieved by integrating Scott Russel linkage to the gripper finger, which translates the horizontal force from the gripper to vertical force to grip the cables (figure 6). 

This finger mechanism consists of four beams, whereby two beams are double the length of the other two. One end of the long beams is attached to the horizontally moving gripper, while their midpoint is attached to the shorter beams. Both of the second end of the long beams are attached to the finger tip, which moves vertically. 

In addition to the beams, this finger design implements an anchor component, which serves as a connection point for the short beams. The anchor component should be stationary and independent from the gripper movement, thus this component is attached to the robot flange. Since this design only has one moving fingertip, the anchor components also serves as second stationary jaw. Furthermore, the anchor component provides railing for the fingertip to ensure that it only moves vertically.

\subsection{Cable Routing}

To enable vertical insertion, the robot grips the cables of industrial plug in a way, that the plug points downward (figure 6). However, this leads to a problem with the routing process. The cables of the industrial plug must be routed through routing forks. Typically, the robot can move above the routing forks while gripping the plug, and move downwards to a position low enough to insert the cable through the clasp element of the forks.  

The routing forks for the industrial plug are located so close to the main board, that the distance between the routing forks and the base of the metal board is less than the height of the plug. Therefore, as the robot moves the connector downwards, the plug will collide with the base of the metal board before the cables are inserted through the routing forks. For this reason, automated cable routing is not possible using the current setup.

\subsection{Plug Insertion}

The socket for the industrial plug has a compliant clasp element, that secures the plug after insertion. Due to this compliant clasp element and the larger size of the plug, the tolerance of the industrial plug is not as tight as the FASTON plug. The plug can still be inserted with a deviation of 1 to 2 mm. As a result, the accuracy of the robot position control is enough to insert the plug directly without the contact-based method from chapter 4.4. 

\section{Experiment Results}
Numerous experiments have been conducted using the proposed cabling method and equipment. The result has shown that with the help the custom gripper fingers and the orientation correction component, the proposed method is able to perform multiple cable assemblies in succession.  This can be seen in the attached video and at \url{https://youtu.be/cv5sMI0DYxM}.

Both of the proposed finger designs have proven to be able to grip the corresponding connector securely, which results in a consistent connector position relative to the robot end effector and subsequently reduce the deviation between the plug and the socket. 

Using this contact-based plug insertion method, the robot can reliably align the plug with the socket, resulting in a robust insertion process. On the other hand, due to the more relaxed tolerance, the second connector type can be inserted to the socket using position control.  

Furthermore, the orientation correction component has functioned as expected, allowing the second end of the FASTON connector to be regripped and inserted.  

Due to geometry constraints, the cable routing process is only performed for the FASTON connector. Using the proposed grip and pull method as well as position control, the robot is able to route the cable to the corresponding points while also manipulating the cable shape to fit among other components on the metal board. 

As reference, the duration of each phases (in seconds) of the cable assembly is listed in the following table. Except for the routing process of the FASTON connector, the robot is able to perform the processes in a reasonable duration. 

\begin{table}[h]
\caption{Duration of Assembly Processes in Seconds}
\begin{tabular*}{\hsize}{@{\extracolsep{\fill}}lllll@{}}
\hline
Connector & Gripping & Insertion & Routing & Orientation Correction\\
\hline
FASTON&   5 &  5& 48 & 9\\
Industrial &9 &  5& -& -\\
\hline
\end{tabular*}
\end{table}

\begin{figure}
    \centering
    \includegraphics[width=\columnwidth]{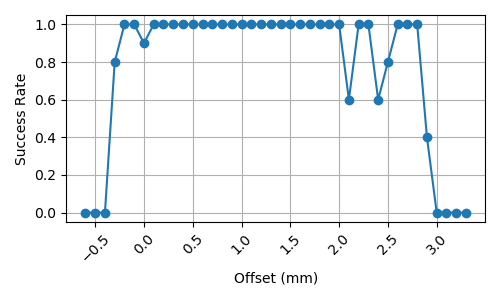}
    \caption{Success rate for over offset in the $y$ direction.}
    \label{fig:success_rate}
\end{figure}

To check the tolerance window realized by the system, we introduce an offset to the goal position of the robot, allowing a controlled error. At each offset, the insertion process is repeated from magazine grasp to plug insertion five times.  The resulting success rate can be seen in Figure \ref{fig:success_rate}, where a $3.0$mm tolerance window has been achieved. The tolerance window seen in Figure \ref{fig:success_rate} is not centered at zero because the taught position is not perfectly centered on the plug. Note that due to the search strategy, the tolerance window in the $x$ direction is large, thus only the $y$ direction is investigated here.

\section{Limitations and Improvement Proposals}
\subsection{Robot Accuracy}
Experiments have shown, the robot is more susceptible to end effector inaccuracy if it is controlled through pose input. Using pose input, the robot can reach the target pose using different configuration of joint angles, resulting in possible kinematic errors.  After switching from pose input to joint angle input, the robot has shown improved end effector accuracy, which also increases the reliability of the cable assembly. 

\subsection{Cable Routing}

Conducted experiments in this paper have proven, that automated cable routing is possible using the proposed method from chapter 4.2. However, the routing process in such a tight workspace requires an intricate teaching process. One reason being that, the robot must be taught a lot of positions to reach the routing points, route the cable, and avoid collision with immediate obstacles. For example, at one point in this routing process, the robot required several positions just to move a couple of centimeters. 

Another reason being that, the taught positions must consider the overall cable shape. The proposed method involves pulling the cable to bring it to the routing points. The positions must be taught in such a way, that the cable is generally bent the correct way. Otherwise the cable can be stuck on obstacle edges, preventing the robot from pulling it further and ultimately failing the routing process. 

For these reasons, the robot teaching for the routing process is time consuming. Furthermore, it takes the robot approximately one minute to perform the routing process, which is relatively a long time, since routing the cable manually for the same process does not take more than five seconds. 

\subsection{Finger Switching System}
While the proposed finger with the Scott-Russel linkage can grip the industrial plug reliably and comply to the geometry of the plug and workspace, it requires an anchor component that is independent from the gripper movement, which is why it is attached to the robot flange. This anchor component will complicate the application of a finger switching mechanism, which allows the robot to assemble different types of connectors in the same cycle.

\subsection{Cable Magazine}
As described in the previous chapter, the magazine utilized within this paper stores the cable plug in a definite position and orientation. However, this magazine can only store one cable at a time, and the cable must be loaded one by one. In an industrial setting, it is preferred to load multiple cables at once to the magazine to save time. 

Thus, an ideal magazine design would be one that fulfill the following requirements: (i) The magazine allows multiple cable to be loaded at once, (ii) The magazine store the cable in a definite position and orientation. Several magazine designs have been proposed within the scope of this paper but none of them can fulfill the two conditions simultaneously.

\subsection{Product Design}
The experiments conducted for this paper have shown, that the feasibility of cable routing using the robot is strongly dependent on the geometry of the workspace. While it is understandable that there are restrictions in product design, the following points can improve automation potential. (i) Making the routing points more accessible to the robot, which will considerably reduce the routing process duration. (ii) Selecting connector type that can be securely gripped with a simple finger design. This will eventually allow for the use a finger switching system that allows multiple connector types to be assembly in one cycle. (iii) Selecting connector type with more relaxed insertion tolerance. This will allow the robot to insert the plug using only position control, foregoing complex insertion method and minimizing cycle time. 

\section{Conclusion}
This paper has proposed novel equipment for cable assembly, namely the fingers designs and the orientation correction device, as well as general cabling method for two types of connectors. Using the proposed equipment and method, the entirety of the FASTON connector assembly can be reliably automated. On the other hand, due to the geometry of the workspace, only gripping and the insertion of the second connector type can be automated. 

By reference to the numerous conducted experiments, multiple challenges for automated cable assembly have been identified. In this regard, multiple limitations are identified and improvement proposals are introduced as reference for future works in automated cable assembly. 

\bibliographystyle{IEEEtran} %

\begin{thebibliography}{00}

\bibitem{trommnau}J. Trommnau, J. Hühnle, J. Siegert, R. Inderka, T. Bauernhasl, "Overview of the State of the Art in the Production Process of Automotive; Wire Harnesses, Current Research and Future Trends," In 52nd CIRP Conference on Manufacturing Systems, 2019. 

\bibitem{yumbla2}F. Yumbla, J. Yi, M. Abayebas, M. Shafiyev, H. Moon, "Tolerance dataset: mating process of plug-in cable connectors for wire harness assembly tasks," 2019.

\bibitem{navas}G.E. Navas-Reacos, D. Romero, J. Stahre, A. Cabalerro-Ruiz, "Wire Harness Assembly Process Supported by Collaborative Robots: Literature Review and Call for RnD," 2022.

\bibitem{hartisch} R. Hartisch, K. Haninger. "High-speed electrical connector assembly by structured compliance in a finray-effect gripper." IEEE/ASME Transactions on Mechatronics, 2023.

\bibitem{haraguchi}Haraguchi et al., "Development of Production Robot System that can Assemble Products with Cable and Connector," 2011.

\bibitem{yumbla}F. Yumbla, M. Abayebas, J. Yi, J. Jeon, H. Moon, "Reposition and Alignment of Cable Connectors Using a Vibration Plate Manipulator for Wire Harness Assembly Tasks," 2021.

\bibitem{yumbla3}F. Yumbla, J. Yi, M. Abayebas, H. Moon, , "Analysis of the Mating Process of Plug-In Cable Connectors for the Cable Harness Assembly Task," 2019.

\bibitem{zhou} Zhou et al., "Design of a Gripper for Cable Assembly with Integrated In-hand Cable Manipulation Functions," 2021. 

\bibitem{heisler}P. Heisler, P. Steinmetz, I.S. Yoo, J. Franke, "Automatization of the Cable-Routing-Process within the Automated Production of Wiring Systems," 2017. 

\bibitem{zhu}J. Zhu, B. Navarro, P. Fraisse, A. Crosnier, A. Cherubini, "Dual-arm Robotic Manipulation of Flexible Cables," In IEEE/RSJ International Conference on Intelligent Robots and Systems (IROS), 2018.

\bibitem{jiao}C. Jiao, X. Jiang, X. Li, Y. Liu, "Vision Based Cable Assembly in Constrained Environment," 2018.

\bibitem{waltersson}G.A. Waltersson, R. Laezza, Y. Karayiannidis, "Planning and Control for Cable-routing with Dual-arm Robot," 2022.

\bibitem{zhang}X. Zhang, C. Li, N. Xi, "Cable Assembly Based on Robot Manipulation and Control," 2022.

\bibitem{sun}B. Sun, F. Chen, H. Sasaki, T. Fukuda,  "Robotic Wiring Harness Assembly System for Fault-tolerant Electric Connectors Mating," 2010.

\end{thebibliography}

\end{document}